\documentclass[10pt,twocolumn]{article}

\usepackage[a4paper,margin=0.75in]{geometry}
\usepackage{graphicx}
\usepackage{caption}
\usepackage{booktabs}
\usepackage{placeins}
\usepackage{amsmath}
\usepackage{times}
\usepackage{placeins}
\usepackage{float}

\title{\textbf{Transformer based Multi-task Fusion Network  for Food Spoilage Detection and Shelf life Forecasting }}

\author{
Mounika Kanulla$^{1}$,
Rajasree Dadigi$^{2}$,
Sailaja Thota$^{3}$,
Vivek Yelleti$^{4}$\\[0.5ex]
$^{*}$SRM University, Andhra Pradesh, India\\
$^{1}$ \texttt{mounika\_kanulla@srmap.edu.in}\\
$^{2}$ \texttt{rajasree\_dadigi@srmap.edu.in}\\
$^{3}$ \texttt{sailaja\_thota@srmap.edu.in}\\
$^{4}$ \texttt{vivek.y@srmap.edu.in}\\
}

\date{}

\begin{document}
\twocolumn[
\maketitle
\begin{abstract}
Food wastage is one of the critical challenges in the agricultural supply chain, and accurate and effective spoilage detection can help to reduce it. Further, it is highly important to forecast the spoilage information. This aids the longevity of the supply chain management in the agriculture field. This motivated us to propose fusion based architectures by combining CNN with LSTM and DeiT transformer for the following multi-tasks simultaneously: (i) vegetable classification, (ii) food spoilage detection, and (iii) shelf life forecasting. We developed a dataset by capturing images of vegetables from their fresh state until they were completely spoiled. From the experimental analysis it is concluded that the proposed fusion architectures  CNN+CNN-LSTM and CNN+DeiT Transformer outperformed several deep learning models such as CNN, VGG16, ResNet50, Capsule Networks, and DeiT Transformers. Overall, CNN + DeiT Transformer yielded F1-score of 0.98 and 0.61 in vegetable classification and spoilage detection respectively and mean squared error (MSE) and symmetric mean absolute percentage error (SMAPE) of 3.58, and 41.66\% respectively in spoilage forecasting. Further, the reliability of the fusion models was validated on noisy images and integrated with LIME to visualize the model decisions. 
\end{abstract}

\vspace{1em}
]

\section{Introduction}

Food wastage is a global challenge that lies at the intersection of economic, environmental, and social concerns. Approximately one-third of all food produced worldwide is wasted annually, with fruits and vegetables contributing significantly due to their perishable nature and susceptibility to rapid quality degradation. Traditional methods of food quality assessment rely mainly on manual inspection, which is time-consuming and prone to human error. With the continuous growth of the global population and increasing demand for food supply, there is a need for automatic, objective, effective, and accurate solutions that can predict the freshness and remaining shelf life of perishable products such as fruits and vegetables. 

Recent developments in artificial intelligence, particularly in the domains of computer vision and deep learning, have demonstrated strong performance in image classification tasks \cite{ref7,ref10}. These models were also applied to agriculture and food quality assessment. However, extant approaches mainly focus on classification tasks, such as identifying the spoilage class of food. These traditional methods fail to predict the remaining shelf life of the food.

The motivation for this work arises from identifying limitations in current food quality assessment systems and recognizing that food quality assessment should be based on multiple factors rather than a single one. A comprehensive understanding of freshness requires the consideration of multiple aspects, including identifying the vegetable type, determining its current spoilage state, and predicting the remaining shelf life. Despite their independence, these tasks deeply interconnect. In real-world deployment, models must also handle challenges such as varying lighting conditions, image quality variations, and sensor noise. The proposed model is designed to handle noisy images with high accuracy. Furthermore, the black-box nature of deep learning models limits their use in critical applications such as food safety. Therefore, the proposed system incorporates multi-output learning, robustness evaluation under noisy conditions, and explainability mechanisms that offer readable perspectives on model decisions.

Current state-of-the-art approaches in food quality assessment mainly rely on single-architecture models such as ResNet, VGG, and MobileNet for classification tasks. Although these methods achieve good accuracy, they suffer from several limitations:

\begin{itemize}

    \item No existing model predicts the remaining shelf life of vegetables.

    \item Most existing work is conducted using publicly available datasets.

    \item Existing models are evaluated only on clean datasets, which limits robustness.

    \item The lack of interpretability makes it difficult for end users to trust the model.

\end{itemize}

The proposed system addresses the above gaps by using novel fusion architectures to create multi-task learning systems capable of vegetable classification, spoilage detection, and shelf life forecasting, while maintaining robustness under noisy conditions and providing understandable predictions through LIME visualization techniques.

\section{Problem statement}
The critical problem addressed in this paper is the development of an effective, automated, robust and interpretable system that not only performs the classification but also predicts  the shelf life of a vegetable. Traditional approaches that rely on human inspection is time-consuming and prone to error. While deep learning  emerged as an alternative solution, the existing models to simulatenously predict all three of them. Further, the model should be able to predict the image even in conditions such as poor 
lighting conditions, low-quality cameras to meet the real-world demands. The architecture should also explain the decision taken by the model.

\section{Novelty Claims}
\begin{itemize}
    \item Predicting the shelf life of the vegetable: \\
    Unlike the extant approaches, the proposed system mainly focuses on predicting the 
    shelf life of the vegetable. As the extant approaches only predict the class of spoilage, 
    the proposed system add up the regression task for that.

    \item Fusion Architectures: \\
    As the single models are unable to generalise and are not very effective, the proposed 
    system fuses the heterogeneous models to improve the robustness and effectiveness.

    \item Systematic Robustness Evaluation: \\
    While most existing research focuses exclusively on the model performance under 
    ideal conditions, this work analyses the robustness of the model by evaluating all the 
    models both on the original clean datasets and synthetically corrupted noisy datasets, 
    providing critical insights into real-world deployment conditions.

    \item Explainability Integration for Multi-output models: \\
    The project implements LIME for a multi-output model, enabling the user to understand which image regions influence different prediction outputs, simultaneously addressing the need for transparency in automated food quality assessment.
\end{itemize}
\section{Literature Review}
Recent advancements in deep learning and computer vision have had a significant impact on research in food freshness and spoilage detection, enabling the assessment of quality using visual information. Most of the existing studies focused on classifying fruits and vegetables into binary or multiple classes using convolutional neural networks (CNNs) and object detection frameworks, which are trained on the publicly available datasets. However, these lack in the regression task, such as shelf-life prediction and also testing the robustness of the models in noisy conditions. These existing works had paved the way for future works and enhancements in the food freshness-related research.

The study \cite{ref1} presents a deep learning-based approach for the evaluation of fruit freshness using YOLOv4 and Tiny YOLOv4 object detection models. As this work mainly focuses on two fruit types, apple and banana, it formulates the task as multi-class classification. The study uses a combination of self-captured and publicly available datasets, and the performance metrics used are accuracy, recall, F1 score. While the approach is nearer to real-time deployment by capturing some of the real-time data, but limited to classification. The article \cite{ref2} presents a YOLOv4-based framework for the classification of fruit and vegetables into multiple classes and also the freshness classification. This study proposes a system that identifies the object category first and then classifies it into fresh or rotten using the Mish activation function. The data augmentation techniques are used to handle variations related to lighting and background. Though the results demonstrate higher average precision compared to YOLOv3 and YOLOv4, it lacks in the regression task. The study \cite{ref3} proposes and fruit freshness classification system using convolutional neural networks and transfer learning based on the AlexNet architecture. This work mainly focuses on data preprocessing steps such as colour uniforming, image resizing, data augmentation and labelling and then followed by fine-tuning of a pre-trained CNN. The model is evaluated using three publicly available datasets. Though it achieves a huge classification accuracy with low computational latency, the approach is limited to single-task classification, mainly focusing on freshness classification.

This work \cite{ref4} proposes an explainable hybrid architecture that combines convolutional neural networks and vision transformers for food image recognition. The model consists of ResNet50 for local feature extraction and a Vision Transformer to capture long-range dependencies using multi-head attention. Compared to existing works, the model demonstrates superior classification performance on multiple diverse datasets. The work also includes the explainability techniques such as Grad-CAM and LIME, which incorporate transparency. The work mainly focuses on single-task food recognition, without addressing freshness grading.
The study \cite{ref5} proposes a transfer learning-based approach for shelf life prediction of mangoes using visual information combined with thermal imaging. The main aim of this work is to estimate the remaining useful life(RUL) of kesar mangoes by classifying the images into 19 classes, ranging from RUL-1 to RUL-18 and no life. Thermal imaging is employed to capture the intrinsic features such as bruises, internal defects and temperature variations and some of the CNN models are evaluated. Though it achieves a high accuracy, it lacks in generalisation to real-world scenarios.
This article \cite{ref6} presents a hybrid convolutional neural network and transfer learning-based approach for assessing the freshness of canned apple fruits. The system is mainly designed to classify the apple image into the three freshness categories – fresh, semi–fresh, and rotten using CNN-based feature extraction. Multiple deep learning architectures are evaluated with the benchmark performance. Though the transfer learning had indicated its effectiveness on both the original and augmented datasets, it fails in the generalizability.
The article \cite{ref7} proposes an ensemble-based multi-task deep learning framework for fruit freshness classification using ResNet-50 and ResNet-101. The model is designed for two classification tasks, one for fruit type identification and another for binary freshness classification. The transfer learning is also employed to handle the limited training data. The evaluation of the model is done using benchmark datasets and also real-world web images, and the model has achieved high classification accuracy.

Reka et al. \cite{ref8} had proposed a deep learning framework for the classification of rotten fruits and the prediction of their shelf life using computer vision techniques. The study included employing CNN-based pre-trained models to classify multiple fruit types into fresh and rotten categories. Among the evaluated models, VGG16 had achieved the highest classification accuracy of 95\% for fruit freshness prediction. In addition, machine learning models such as Gaussian Naïve Bayes and Random Forest are used for shelf-life prediction. This work demonstrates the effectiveness of combining deep learning and machine learning models.
Zhang et al. \cite{ref9} proposed a fruit freshness detection system based on a multi-task learning convolutional neural network, which performs the multi tasks, freshness detection and fruit type classification. The model employs CNN for feature extraction and two task-specific fully connected heads to capture the semantic correlations between the tasks. The results are analysed on a publicly available fruit image dataset, where MTL approach outperforms single-task learning, achieving 93.24
Khan et al. \cite{ref10} had proposed an enhanced attention-based convolutional neural network with the combination of explainable AI to enhance the accuracy of fruit and vegetable classification and user interpretability. This study addresses the limitations of the traditional CNN models, such as computational complexity and lack of interpretability, by introducing a customised pooling strategy. The model is evaluated on a dataset that contains 141 classes. EA-CNN had achieved an accuracy of 98.1\% with the least number of training iterations. The model was also evaluated on a real-world Fruit Recognition dataset, achieving 96\% accuracy, proving its generalisation.

Putra and the team \cite{ref11} did a study on the problem of food wastage. They had applied deep learning models for automatic freshness classification of fruits and vegetables. They had used the Kaggle dataset, which contains fresh and rotten data. Mainly focused on binary classification. They have tested 3 CNN models that we already made – MobileNetV2, VGG19, and EfficientNetV2S. Their results show that EfficientV2S achieves the best accuracy and performance. It shows that CNNs, which are light but strong, are fast and good for checking food quality. This work only does one type of classification and does not try to predict shelf life or handle multiple outputs. Also, it uses a public dataset and does not check how solid it is or explain how the model makes its decisions. They \cite{ref12} had worked on deep learning that can be used to automatically detect produce spoilage. The goal is to reduce food wastage in homes and retail.  They had developed their own CNN and compared its effectiveness with the most commonly used transfer learning models for identifying whether they are fresh or rotten. CNN achieves high accuracy, and maintaining computational complexity is low. They had only used binary classification and did not consider freshness levels or shelf-life estimation. They \cite{ref13} had worked on a fruit freshness grading system by using deep learning and analysing images. They had used models like AlexNet, VGG, ResNet, and GoogleNet. Also used YOLO to focus on certain parts of the image helps them to get more effective grades. This method is good at sorting fruits by freshness, but it only classifies them. It does not predict the shelf life or explain how the model makes its decisions.This study \cite{ref14} proposes an automated fruit freshness monitoring approach using a ResNet-101 model. It is enhanced with a non-local attention mechanism helps to find out small changes in the fruit appearance. By adding this attention part, the model is good at finding bad spots.
This test showed that this way works better than CNN models like ResNet-50 and VGG-16 in accuracy. But it only checks freshness and does not predict shelf life.

This study \cite{ref15} is about a CNN model that uses deep learning for automatic spoilage of tomatoes using a self-prepared image dataset. The model divides tomatoes into two groups: good to eat or spoiled. They had achieved very high accuracy by using hyperparameter tuning. Evaluation metrics are the confusion matrix and Pearson’s correlation, and comparing it to how people decided the tomatoes by using sight and smell. But it only works for tomatoes, and can only tell if they are good or bad. It can’t predict the shelf life, handle blurry images well, or do more than one thing at a time. The system \cite{ref16} can automatically check how fresh fruits and vegetables are. It not only predicts the freshness or spoilage, but also predicts that the fruit or vegetable has 3 classes: pure-fresh, medium-fresh, or rotten.  They had used VGG-16 to classify freshness and YOLO to find the fruits and vegetables in the image, so it knows what it’s looking at and where it is. The authors created a big dataset with 60,000 pictures of 11 kinds of produce and made it available for others to use. It is pretty accurate and can even run on an Android app in real-time. But it doesn’t predict the shelf-life of fruits or vegetables. This study \cite{ref17} proposes an automated system for detecting the freshness of fruits and vegetables. Used deep features extracted from multiple pre-trained CNN models. Used deep features from GoogLeNet, DenseNet-201, and ResNeXt-101. They are fused and reduced using PCA before classification with regular machine learning methods. They had achieved high accuracy by using multi-model feature fusion for freshness detection compared to single-model approaches. However, the technique focuses only on freshness classification and does not predict the shelf-life, end-to-end deep learning or explainability mechanisms.

This study \cite{ref18} analyses how well transfer learning-based CNNs and residual CNN models perform for fruit freshness classification using an image dataset. Multiple regular CNN models and residual networks are evaluated to identify fruit type and how fresh it is. Residual networks did the best and got over 99\% accuracy at predicting freshness in the dataset.  But it only classifies, and does not predict the shelf-life of a fruit.This \cite{ref19} is a deep learning-based system for the classification of fruits and vegetables, also with ripeness assessment. They had used two MobileNetV2 models, one model for identifying the fruit or vegetable and another for determining the level of ripeness for the classes that they had selected.  Both the classification and prediction of ripeness are achieved with high accuracy in CNN. This method evaluates only ripeness and a particular set of classes. It does not predict spoilage and shelf-life. The main focus of the research \cite{ref20} is automated quality analysis for apples by using semantic segmentation on rotten regions on the fruit surface using deep learning. The authors have come up with 2 different models, namely UNet and Enhanced UNet, or En-UNet, which have been effective for defect segmentation for images of apple peel. En-UNet performs the better than regular UNet in tests, getting better accuracy and mean IoU scores, which means it could be good for predicting defects quickly. But right now, it only segments and checks if fruit is fresh or not. It can’t divide the things into different freshness levels or predict the shelf-life. This \cite{ref21} is a real-time quality classification system. The goal is to reduce manual labor in consumer markets. They had used YOLO-V3 for fruit detection and continuous tracking to check the external quality of round fruits like apples and oranges by using visual features like size and shape. The system focuses only on external quality detection and does not consider freshness, shelf-life prediction, or multi-level spoilage analysis.They \cite{ref22} had created a system that is to check the freshness of apples by taking pictures and classifying them into fresh and rotten. Used computer vision techniques. Primarily, it divides the apple from the background in the pictures. Then it uses PCA to select the best features, and after that looks at particular details in the image. They had used machine learning (SVM and KNN) to check whether the apple is fresh or rotten. SVM has worked well for this. They had used only binary classification, whether the apple is good or bad. 

This \cite{ref23} is an IoT-enabled hybrid deep learning system that automatically identifies fresh fruits and vegetables. The goal is to improve efficiency in agriculture and retail. They had combined EfficientNetB7 and ResNet50 models to bring together the best features from both approaches to get highly accurate results across multiple datasets.  The proposed hybrid model performs better than individual CNN models and provides faster responses, and it is suitable for real-time use. However, it mainly focuses on freshness classification and identification. They do not predict spoilage stages and shelf-life of fruits and vegetables. They \cite{ref24} had proposed a hybrid deep learning model that combines CNN and bidirectional LSTM layers. They have the disadvantage of manual feature extraction. That’s why they had proposed this model. By combining these models, the researchers successfully captured both the visual and spatial details, which leads to more accurate freshness results of fruits and vegetables. Though the method focuses only on freshness classification, it does not extend to shelf-life estimation. This paper \cite{ref25} presents Freshcheck, a two-stage deep learning model designed for the freshness assessment of fruits and vegetables using visual features. They had used MobileNetV2 for the classification divides into 3 levels: fresh, medium fresh, and not fresh. Using transfer learning along with a regression-to-classification approach allows the system to accurately predict freshness levels. However, the approach does not include the shelf-life prediction and performance analysis under noisy conditions.

\section{Proposed Methodology}

\subsection{Proposed Model -- 1: MobileNetV2 and DeiT Transformer Fusion Model}

The architecture is the combination of MobileNetV2 and the DeiT transformer for the multi output model, which can predict the three outputs: vegetable type, spoilage class and the  shelf life of the vegetable. In this specific architecture, the MobileNetV2 model is used for the classification task and the DeiT Transformer for the regression task. The MobileNetV2  model captures the local spatial features, such as colour differences, patches on the vegetable, which are useful for the classification and the DeiT transformer is used to capture the global features and identify high-level structural patterns such as shape irregularities, overall degradation.
After the features are extracted, the predictions from the MobileNetV2 and DeiT are combined through the fusion mechanism, which is simple concatenation. The main advantage of this architecture is the fusion technique, where the model combines the strengths of both models. This fusion model predicts the three outputs: vegetable type, spoilage class and the shelf life of the vegetable. Then the LIME explainability technique has been added to enhance the trust and interpretability for the user. LIME generates region-based explanations to show which parts of the image contributed for the major part of the model's decision-making. Finally, the entire model had been deployed using the Flask API framework, which improves the model's usability. Users can upload the images, and the model will predict the three outputs: the type of the vegetable, the spoilage class, and the shelf life of the vegetable, making the system more transparent and usable for real world problems.

\begin{figure*}
    \centering
    \includegraphics[width=1\linewidth,height=0.25\textheight,keepaspectratio=false]{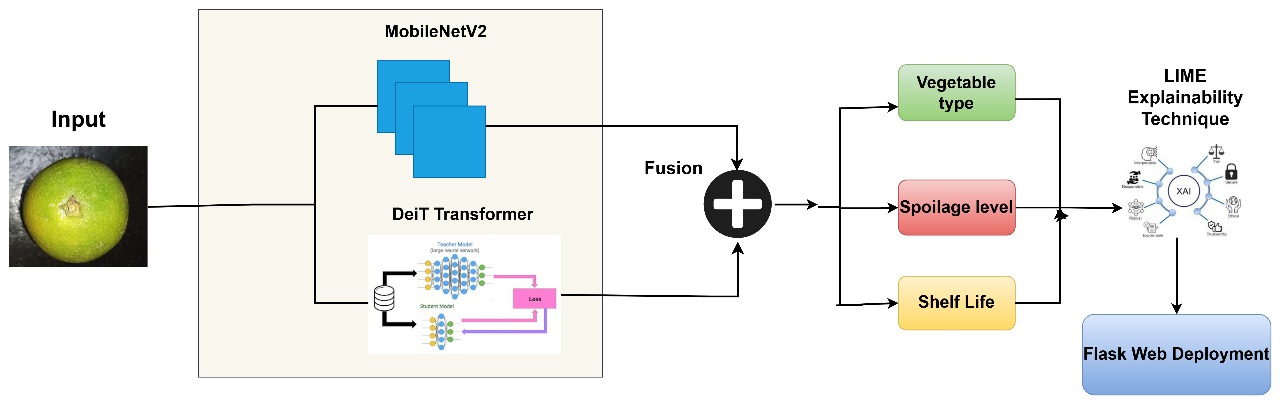}
    \caption{Proposed Methodology-1: MobileNetV2 + DeiT tranformer}
    \label{fig:prop-1}
\end{figure*}

\begin{figure*}
    \centering
    \includegraphics[width=1\linewidth,height=0.25\textheight,keepaspectratio=false]{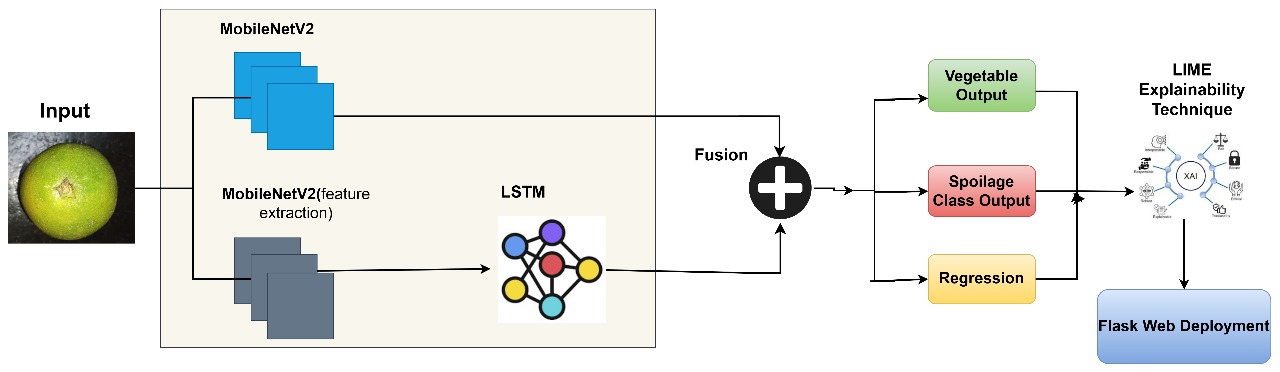}
    \caption{Proposed Methodology-2: MobileNetv2 + MobileNetv2\_LSTM fusion}
    \label{fig:prop-2}
\end{figure*}
\subsection{Proposed Model -- 2: MobileNetV2 and MobileNetV2 -- LSTM Fusion}

This architecture is the combination of MobileNetV2 and MobileNetV2 -- LSTM for the multi-output model, which supports the three outputs. The MobileNetV2 model is used for the classification technique, and the combination of MobileNetV2 and LSTM is used for the regression task. For the classification task, the MobileNetV2 was used for feature extraction and also for the classification of vegetable type and spoilage class. For the regression task, the MobileNetV2 model is used for feature extraction, and the LSTM is used for the regression task, predicting the shelf life of the vegetable.

Once the features are extracted by MobileNetV2, the classification output from MobileNetV2 and the regression output from the LSTM are fused using simple concatenation. The fused feature vector is fed into three separate output heads. This step merges the strengths of both models, especially in the regression task as the feature extraction is done by MobileNetV2 and the regression task, like capturing the numerical patterns, is done by LSTM. To enhance the trust and interpretability, the LIME explainability technique has been added to the model. The LIME highlights the part of the image that is useful for contributing to the decision-making for the model, which helps the user to understand why a model is classified as fresh or spoiled.

\subsection{Explainability for the proposed Model}
The LIME is applied to the MobileNetV2 and DeiT fusion model and MobileNetV2 and MobileNetV2 -- LSTM fusion model to enhance the trust and transparency for the users. This is used to provide the explanations for the predictions generated by the multi-output architecture. As this fusion model integrates many complex operations, interpreting the decision-making of the system is difficult. This problem can be overcome by the LIME by identifying the particular regions of the image that strongly contributed to the model's decision-making in the three outputs, like vegetable classification, spoilage type classification, and shelf life of the vegetable.

\begin{figure}[htbp]
    \centering
    \includegraphics[width=\linewidth]{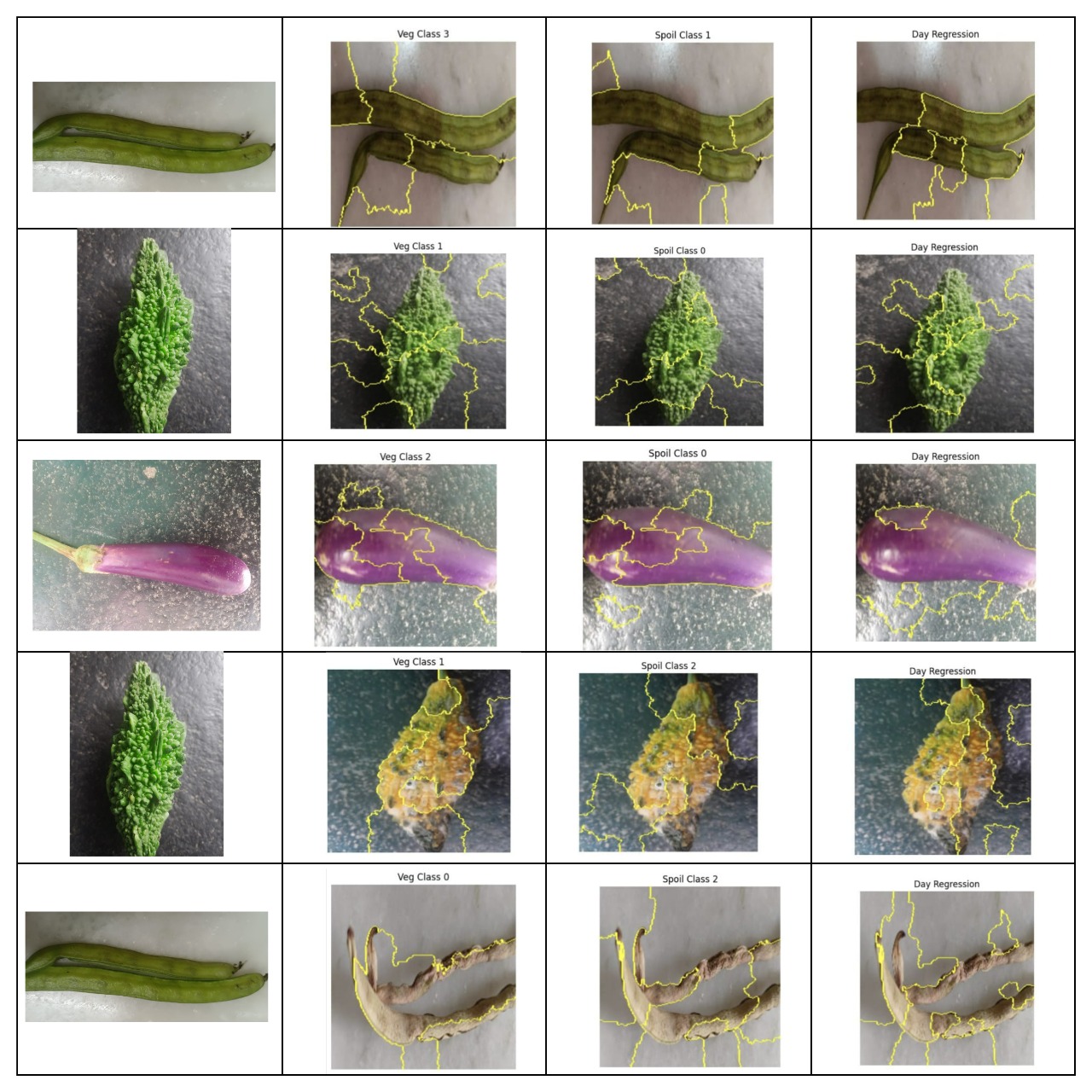}
    \caption{LIME--based explainability maps highlighting the regions influencing vegetable class prediction, spoilage detection, and day-wise freshness regression in the proposed model 1.}
    \label{fig:explainability}
\end{figure}

\begin{figure}[htbp]
    \centering
    \includegraphics[width=\linewidth]{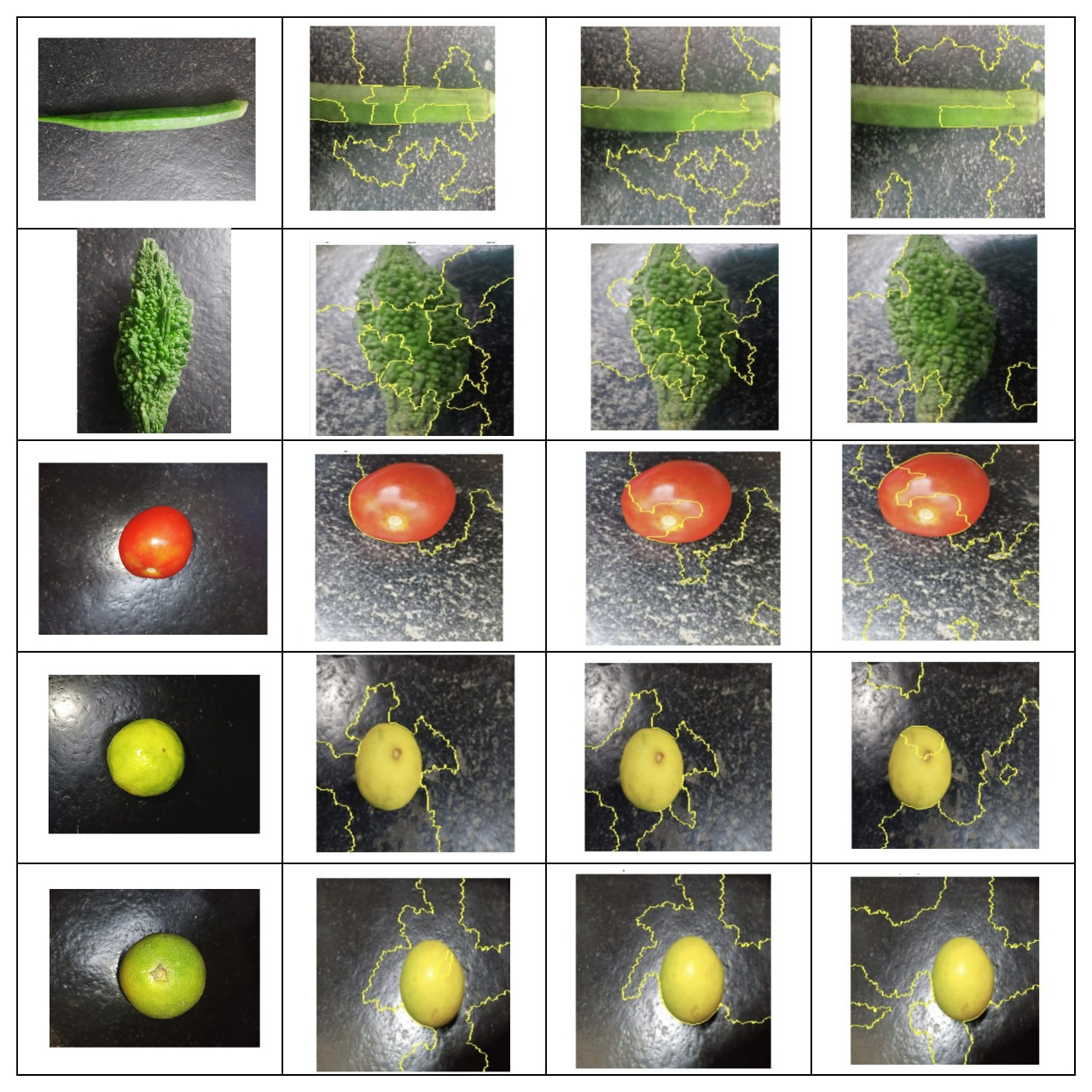}
    \caption{LIME--based explainability maps highlighting the regions influencing vegetable class prediction, spoilage detection, and day-wise freshness regression in the proposed model 2.}
    \label{fig:explainability}
\end{figure}

\section{Dataset Description}
The dataset used in this work is a constructed resource which is mainly designed for 
working with 3 specific outputs: vegetable classification, spoilage class prediction and also 
shelf-life prediction of a vegetable/fruit. The dataset consists of good-quality images of 
various vegetable types such as mosambi, lemon, tomato, bitter guard, ladies finger, brinjal, 
green beans, and beans, which are captured with a high resolution. Each image in the 
dataset is associated with three different labels that provide a complete characterisation of 
vegetable type, a spoilage class indicating the freshness of the vegetable, and the day 
number on which it was captured.

\begin{figure}[htbp]
    \centering
    \includegraphics[width=\linewidth]{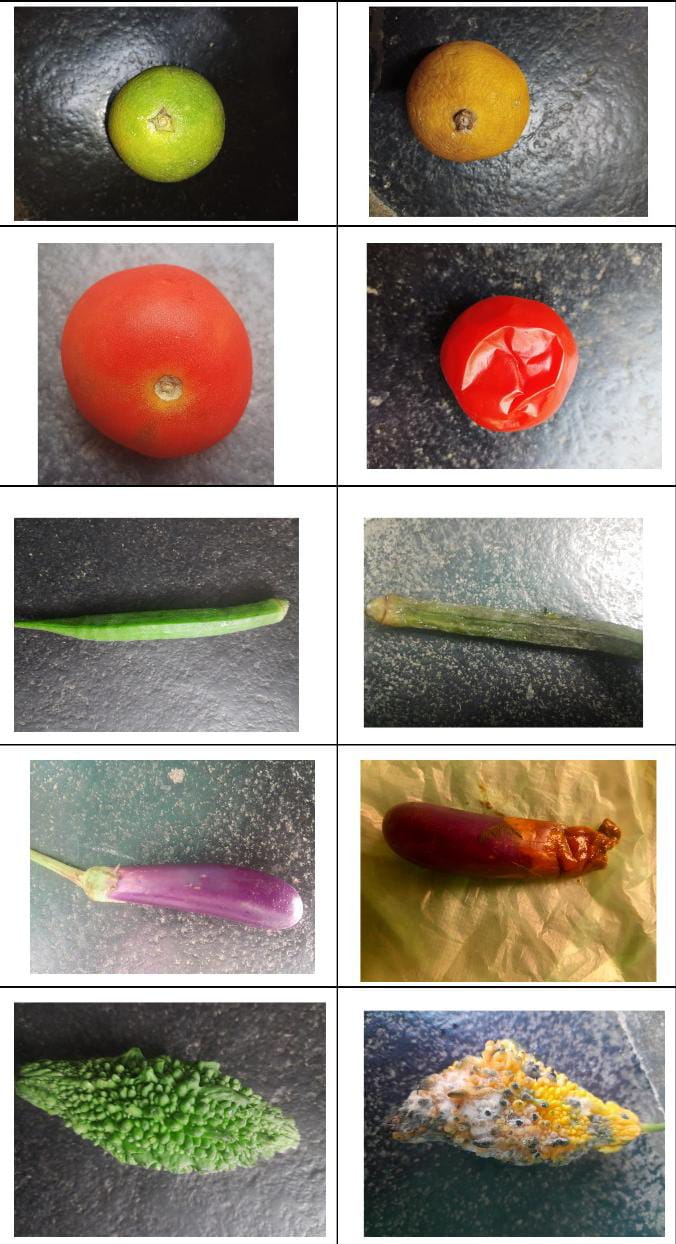}
    \caption{Sample images from the self-constructed dataset illustrating fresh and spoiled conditions across different fruits and vegetables.}
    \label{fig:dataset_samples}
\end{figure}

The data collection involved capturing the images daily until they were completely spoiled. 
Primarily, some random vegetables and fruits are chosen to construct the dataset. The set 
of vegetables is observed continuously until they are spoiled. Six images are captured every 
day for any type of vegetable to maintain uniformity and class balance. To continuously 
track the spoilage of the vegetables, the images are captured at a specific time gap, 
especially morning, afternoon and evening, such that each one has six images in all 
directions. 
\begin{figure}[htbp]
    \centering
    \includegraphics[width=\linewidth]{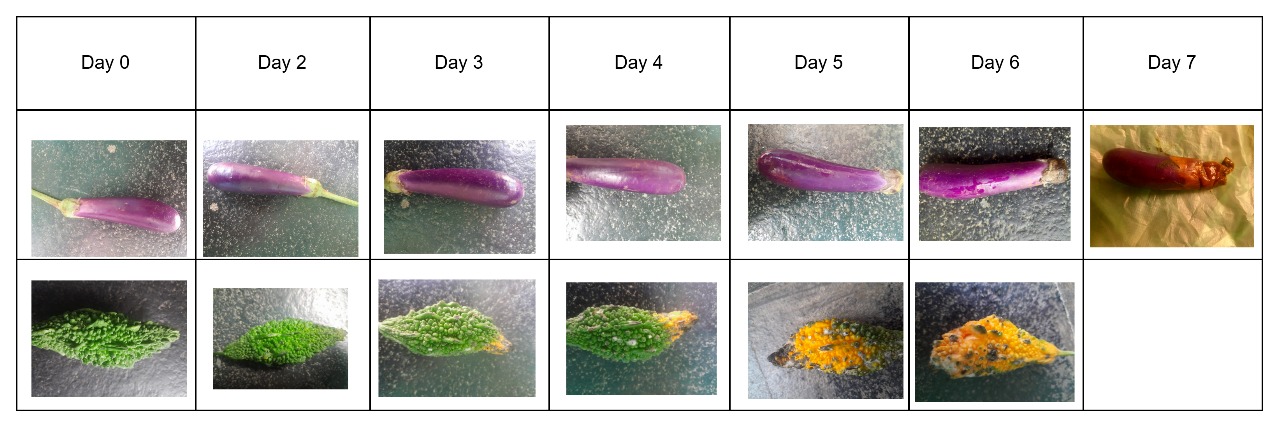}
    \caption{Day-wise visual progression of spoilage in vegetables captured under controlled conditions for freshness regression analysis.}
    \label{fig:daywise_spoilage}
\end{figure}

The dataset was labelled according to the three classes. There are eight classes specifying 
the type of vegetable. Coming to the spoilage classes, there are three classes: mainly fresh, 
slightly spoiled and completely spoiled. And the number is also associated with the label 
itself, stating the day number on which it is captured. The dataset is organised in such a 
way that the 3 labels can be specified. The main folder is the dataset, which includes eight 
subfolders labelled with vegetable names. These folders are further divided into subfolders 
representing the day number and spoilage class. Each day folder contains 6 images.

In the folder name dayx\_y, x represents the day number on which it was captured, and y 
represents the spoilage class (1- fresh, 2-slightly spoiled, 3-completely spoiled).

The noisy dataset was created using the original dataset. In the dataset, in each day folder, 
The random noise was introduced for two images to ensure 25\% of the images are noisy.

\section{Implementation}
\subsection{Classification and Fusion Models}

\textbf{A. MobileNetV2} \\
MobileNetV2 \cite{ref26} is a lightweight convolutional neural network that is mainly designed for 
feature extraction from images, so this is used as the main baseline model in this work. The 
model consists of inverted residual blocks and depth wise separable convolutions to reduce 
the computational cost. These advantages are the main reason for the efficiency of this 
model. In this work, it is used to predict the three output heads.

\textbf{B. MobileNetV2+MobileNetV2\_LSTM \cite{ref27}} \\
In this architecture , the MobileNetV2 model is used for the classification task and the 
LSTM for the regression task. For the regression task also the MobileNetV2 model is used 
for feature extraction and LSTM for capturing the temporal-like dependencies. Both 
models are fused using the simple concatenation technique.

\textbf{C. VGG16} \\
This model \cite{ref28} is a deep CNN architecture that uses multiple 3x3 convolutional layers to learn 
the hierarchical features of the images. The main advantage of the VGG16 is its depth 
making it able to capture the detailed spatial information.

\textbf{D. ResNet50} \\
This architecture \cite{ref29} is mainly useful to solve the problem of vanishing gradients. Its skip 
connections are useful for the model to learn the simple and complex features. This model 
is mainly used to extract the deep spatial features of an image.

\textbf{E. MobileNetV2+VGG16 Fusion Model \cite{ref30}} \\
This fusion model combines the advantages of both the MobileNetV2 model and the 
VGG16 model. The MobileNetV2 model is used for the classification head, and the 
VGG16 is used for the regression. Both these models are fused using simple concatenation.

\textbf{F. MobileNetV2+ResNet50 Fusion Model \cite{ref31}} \\
This fusion model highlights the capabilities of both models. This model combines the 
strengths, such as the powerful residual learning capability of ResNet50 and the lightweight 
design of MobileNetV2. In this architecture, the MobileNetV2 is used as the classification 
head and ResNet50 as the regression head.

\textbf{G. Capsule Networks} \\
This model \cite{ref32} represents the features as vectors, which allows the model to capture the 
hierarchical features. This model will be efficient when the images require structural 
understanding.

\textbf{H. MobileNetV2+Capsule Network Fusion Model \cite{ref33}} \\
In this architecture, the MobileNetV2 model is used for classification purposes, whereas 
the capsule network model is used for a regression task, where it groups the features into 
capsules, mainly focusing on the orientation of features.

\textbf{I. DeiT (Data-efficient Vision Transformer)} \\
DeiT \cite{ref34} is a transformer model that focuses on global relationships rather than the local spatial 
features. It does not use any convolution techniques such as CNNs. It works efficiently 
with the smaller training datasets.

\begin{table*}[!htbp]
\centering
\caption{Model-wise Hyperparameter Configuration}
\label{tab:hyperparams1}
\renewcommand{\arraystretch}{1.3}
\begin{tabular}{|p{5cm}|p{12cm}|}
\hline
\textbf{Model} & \textbf{Hyperparameters} \\ \hline

MobileNetV2 &
Learning rate: [$1e^{-4}$] \\
& Epochs: [25] \\
& Activation: ['softmax', 'softmax', 'linear'] \\
& Optimizer: ['Adam'] \\
& Dropout: [0.3] \\
& Loss Functions: ['sparse\_categorical\_crossentropy','sparse\_categorical\_crossentropy', 'mse'] \\ \hline

MobileNetV2+MobileNetV2-LSTM (Fusion) &
Both CNN Classifier and CNN-LSTM Regressor \\
& Learning rate: [$1e^{-4}$] \\
& Epochs: [25] \\
& Activation: ['softmax', 'softmax', 'linear' \\
& Optimizer: ['Adam'] \\
& Dropout: [0.3] \\
& Loss Functions: ['categorical\_crossentropy', 'categorical\_crossentropy', 'mse' \\ \hline

VGG16 &
Learning rate: [$1e^{-4}$] \\
& Epochs: [10] \\
& Activation: ['relu', 'softmax', 'softmax', 'linear' \\
& Optimizer: ['Adam'] \\
& Dropout: [0.3] \\
& Loss Functions: ['sparse\_categorical\_crossentropy','sparse\_categorical\_crossentropy', 'mse' \\
& Base Network: VGG16 (weights = "imageNet", include\_top=False, frozen) \\ \hline

ResNet50 &
Learning rate: [$1e^{-4}$] \\
& Epochs: [10] \\
& Activation: ['relu', 'softmax', 'softmax', 'linear' \\
& Optimizer: ['Adam'] \\
& Dropout: [0.3] \\
& Loss Functions: ['sparse\_categorical\_crossentropy','sparse\_categorical\_crossentropy', 'mse'] \\
& Base Network: VGG16 (weights = "imagenet", include\_top=False, frozen) \\ \hline

MobileNetV2 + VGG16 (Fusion) &
Learning rate: [$1e^{-4}$] \\
& Epochs: [15] \\
& Activation: ['relu', 'softmax', 'softmax', 'linear'] \\
& Optimizer: ['Adam'] \\
& Dropout: [None] \\ 
& Loss Functions: ['categorical\_crossentropy', 'categorical\_crossentropy', 'mse'] \\
& Base Network: VGG16 (weights = "imagenet", include\_top=False, frozen)\\ \hline

MobileNetV2 + Resnet50(Fusion) &
Learning rate: [$1e^{-4}$] \\
& Epochs: [Not specified] \\
& Activation: ['relu', 'softmax', 'softmax', 'linear'] \\
& Optimizer: ['Adam'] \\
& Dropout: [0.3] \\
& Loss Functions: ['sparse\_categorical\_crossentropy','sparse\_categorical\_crossentropy', 'mse'] \\ \hline

\end{tabular}
\end{table*}

\begin{table*}[!htbp]
\centering
\caption{Model-wise Hyperparameter Configuration}
\label{tab:hyperparams1}
\renewcommand{\arraystretch}{1.3}
\begin{tabular}{|p{5cm}|p{12cm}|}
\hline
\textbf{Model} & \textbf{Hyperparameters} \\ \hline

Capsule Networks &
Learning rate: [$1e^{-3}$] \\
& Epochs: [10] \\
& Activation: ['relu', 'softmax', 'softmax', 'linear'] \\
& Optimizer: ['Adam'] \\
& Dropout: [0.3] \\
& Loss Functions: ['sparse\_categorical\_crossentropy','sparse\_categorical\_crossentropy', 'mse'] \\ \hline

MobileNetV2 + Capsule Networks &
Learning rate: [Default Adam $\sim 1e^{-3}$] \\
& Epochs: [10] \\
& Activation: ['relu', 'softmax', 'softmax', 'linear' \\
& Optimizer: ['Adam'] \\
& Dropout: [0.3] \\
& Loss Functions: ['sparse\_categorical\_crossentropy','sparse\_categorical\_crossentropy', 'mse']\\
\hline

DeiT Transformer &
Learning rate: [$1e^{-4}$] \\
& Epochs: [5] \\
& Activation: ['relu', 'softmax', 'softmax', 'linear' \\
& Optimizer: ['Adam'] \\
& Dropout: [None] \\
& Loss Functions: ['cross\_entropy','cross\_entropy', 'mse']\\
\hline

MobileNetV2 + DeiT Transformer (Fusion) &
Learning rate: [$1e^{-4}$] \\
& Epochs: [5] \\
& Activation: ['relu', 'softmax', 'softmax', 'linear' \\
& Optimizer: ['Adam'] \\
& Dropout: [None] \\
& Loss Functions: ['cross\_entropy','cross\_entropy', 'mse'] \\
& Base Network: VGG16 (weights = "imagenet", include\_top=False, frozen) \\ \hline

\end{tabular}
\end{table*}

\textbf{J. MobileNetV2 + DeiT Fusion Model \cite{ref35}} \\
The fusion model pairs MobileNetV2's simple localized feature extraction and DeiT's 
global feature extraction. In this architecture, the MobileNetV2 model is used for 
classification heads and the DeiT transformer is used for the regression head. This hybrid 
design had combined the strengths of the two models.

\section{Workflow}
\subsection{Model Workflow}

\textbf{Initial Phase:}

\textbf{Step--1:} \\
The implementation starts with loading and processing the dataset, which can support three 
outputs: the type of the vegetable, the spoilage class, and the shelf life prediction of the 
vegetable. The data is processed by dividing the dataset into train, test and validation 
datasets. The images are also resized, and labels are encoded.

\textbf{Step--2:} \\
The feature extraction is performed using the Convolutional Neural Networks. The 
MobileNetV2 is used to capture localised spatial features, and the DeiT transformer is used 
for capturing the global spatial features.

\begin{figure}[htbp]
    \centering
    \includegraphics[width=0.55\linewidth]{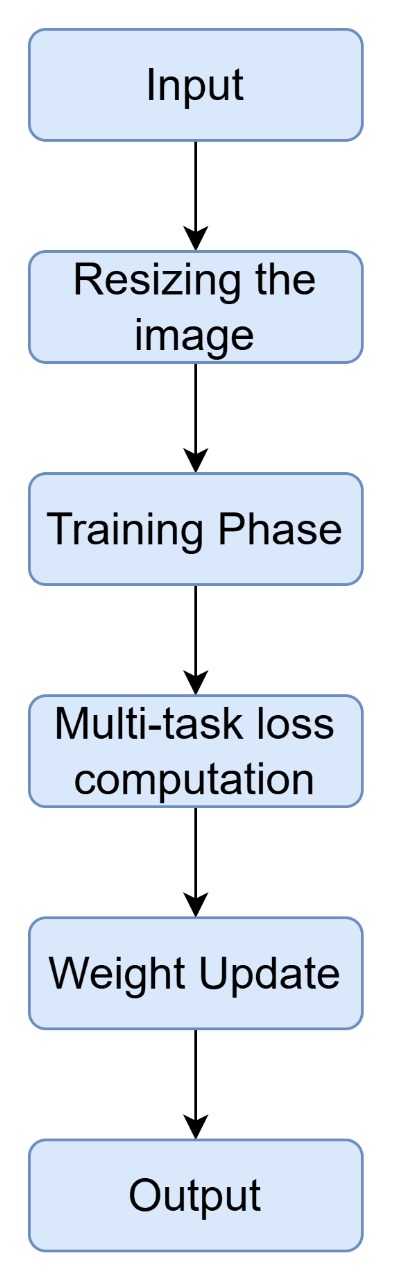}
    \caption{Workflow of the proposed multi-task deep learning framework for freshness classification and regression.}
    \label{fig:workflow}
\end{figure}

\textbf{Training Phase:}

\textbf{Step--3:} \\
After completion of the feature extraction, the models are designed to predict the three 
outputs, such as the type of the vegetable, the spoilage class, and the shelf life of the 
vegetable. The primary model used in the fusion technique is designed for the classification 
task, and the secondary model is designed for the regression task.

\textbf{Step--4:} \\
The fusion is performed for the two models which are designed for the multi output. The 
fusion technique used here is simple concatenation. This technique has achieved the highest 
accuracy where it got more strength by combining the two models.

\textbf{Testing Phase:}

\textbf{Step--5:} \\
The final step includes testing the model using appropriate metrics. The metrics used for 
classification is accuracy and F1 score. The metrics used for regression is MSE and 
SMAPE. These metrics helps to evaluate how well the model is being effective on the each 
output of the dataset.

\section{Results And Discussions}
\subsection{Performance of models on the original dataset}

This section describes the evaluation of various deep learning architectures on the original 
dataset and also the limitations of different neural networks. The performance of these 
models is mainly based on 4 key metrics that characterise models' effectiveness for three 
prediction objectives: Vegetable F1 score for measuring the accuracy of vegetable type 
classification, spoilage F1 score for measuring the accuracy of spoilage detection, Mean 
Squared Error(MSE) is used to test the days-to-spoil regression prediction where lower 
value suggests that better performance of the model and Symmetric Mean Absolute 
Percentage Error(SMAPE) provides a percentage based measure where lower percentage 
represent superior capability. 

The baseline MobileNetV2 architecture achieves a good amount of F1 score of 0.84, which 
indicates strong capability in extracting the features of vegetables and can classify them 
strongly. Coming to the spoilage F1 score, it showed a moderate performance with 0.59, 
specifying that it is unable to extract the exact states of the spoilage class. It records MSE 
of 32.7 and SMAPE of 86.90\% indicating a huge error in predicting the shelf life. This is 
because MobileNetV2 is lightweight and suitable for simple classification tasks, but fails 
in spoilage classification because it requires capturing more detailed features. The 
MobilenetV2-LSTM fusion architecture was the first fusion model implemented, which 
combines the efficient feature extraction and LSTM’s temporal modelling capabilities to 
capture the progressive nature of the vegetable spoilage. This architecture achieves a 
remarkable vegetable F1 score of 0.97, which has increased from the previous standalone 
model. The spoilage F1 score of this model remains unchanged from the previous baseline 
model. There is a significant improvement in the regression task, where the model reduces 
the error to 5.07 and SMAPE to 42.07\%. This exceptional performance is due to 
incorporating the LSTM into the fusion model, which highlights the ability of LSTM in 
predicting the sequential patterns, which has enabled the more accurate prediction of time
based deterioration.  
\begin{table*}[t]
\centering
\caption{Performance comparison of models on the original dataset}
\label{tab:performance}
\renewcommand{\arraystretch}{1.4}
\begin{tabular}{|l|c|c|c|c|}
\hline
\textbf{Model} & \textbf{Vegetable F1 Score} & \textbf{Spoilage F1 Score} & \textbf{MSE} & \textbf{SMAPE(\%)} \\
\hline
MobileNetV2 & 0.84 & 0.59 & 32.7 & 86.90 \\
\hline
MobileNetV2+MobileNETV2 LSTM(Fusion) & 0.97 & 0.58 & 5.07 & 42.70 \\
\hline
VGG16 & 0.67 & 0.5 & 14.8 & 65.38 \\
\hline
Resnet 50 & 0.07 & 0.34 & 18.04 & 31.60 \\
\hline
MobileNetV2+VGG16 (Fusion) & 0.41 & 0.36 & 18.44 & 63.75 \\
\hline
MobileNetV2+Resnet50 (Fusion) & 0.87 & 0.64 & 31.74 & 82.59 \\
\hline
Capsule Networks & 0.04 & 0.23 & 22.72 & 68.34 \\
\hline
MobileNetV2+Capsule Networks & 0.07 & 0.22 & 15.25 & 78.81 \\
\hline
DeiT Transformer & 0.98 & 0.65 & 18.67 & 65.40 \\
\hline
MobileNetV2+DieT Transformer(Fusion) & 0.98 & 0.61 & 3.58 & 41.66 \\
\hline
\end{tabular}
\end{table*}

The VGG16 architecture, though a much deeper and complex model, it had only moderate 
results by achieving a Vegetable F1 score of only 0.67, unable to classify even the type of 
the vegetable. Even the spoilage F1 score is 0.50, which is less than the MobilenetV2 
baseline. This is because VGG16 is a large and heavy model, where there is a chance of 
lacking feature reuse when working with small datasets. But coming to MSE, it performed 
better than the baseline model, but comparatively less than the fusion model, with 14.8 and 
15 
an SMAPE of 65.38\%. ResNet50 is nearly unable to predict the type of the vegetable and 
a spoilage F1 score of 0.34, poor, but is unable to surpass MobileNetV2. The regression 
performance shows an MSE of 18.04, underperforming when compared with the fusion 
model. The MobileNetV2 and VGG16 fusion architecture showed a poor vegetable F1 
score of 0.41, the score had reduced from the baseline model but the spoilage F1 score 
remains low. Though the model improves in classification, it lacks in generalizability in 
regression.

The MobileNetV2 and ResNet50 fusion architecture performed very well on classifying 
the vegetable type with an F1 score of 0.87, whereas the CNN had compensated the 
weakness of ResNet50 baseline model. But the spoilage F1 score is 0.64, showing that 
ResNet50 had contributed towards spoilage-related texture analysis. The regression error 
is very high, having an MSE of 31.74 and SMAPE of 82.59\%. Capsule Networks perform 
poorly across all outputs as they require large datasets to generalise the spatial hierarchies, 
and the regression capability was very low. Though the capsule network is fused with 
MobileNetV2, it has not improved the performance, as the capsule features remain too 
unstable for fusion.  

The DeiT transformer had achieved an excellent vegetable F1 score of 0.98 and even a 
good spoilage F1 score, as the transformers are good at capturing the fine texture variations. 
But the transformer is unable to outperform the fusion technique in MSE and SMAPE. The 
fusion model of MobileNetV2 and Deit Transformer had achieved the overall best 
performance in both the classification and regression metrics. It had achieved a Vegetable 
F1 score of 0.98, a Spoilage F1 score of 0.61, MSE of 3.58 and, coming to SMAPE, it 
achieved 41.66\%. While MobileNetV2 works well for image classification, Deit 
transformer has worked well for regression, which made it one of the strongest models. 
Fusion models had outperformed the baseline models because they combine the strengths 
of different architectures. The MobileNetV2 is excellent at extracting the spatial features, 
while models such as LSTM and transformers have worked for regression. This fusion 
architecture had worked well for the multi-output model, especially when concatenated 
with different types of models. The fusion models had achieved higher F1 scores by 
learning the patterns in the images for vegetable classification and spoilage level 
classification, and they reduced MSE/SMAPE by capturing numerical relationships. 

\subsection{Performance of the models on noisy dataset}
This section describes the evaluation of various deep learning architectures on the noisy 
dataset and also the limitations of different neural networks while testing the robustness. 
The performance of these models is mainly based on 4 key metrics that characterise models' 
effectiveness for three prediction objectives: Vegetable F1 score for measuring the 
accuracy of vegetable type classification, spoilage F1 score for measuring the accuracy of 
spoilage detection, MSE and the SMAPE for regression.

\begin{table*}[t]
\centering
\caption{Performance comparison of models on the noisy dataset}
\label{tab:performance_noisy}

\renewcommand{\arraystretch}{1.4}

\begin{tabular}{|l|c|c|c|c|}
\hline
\textbf{Model} & \textbf{Vegetable F1 Score} & \textbf{Spoilage F1 Score} & \textbf{MSE} & \textbf{SMAPE(\%)} \\
\hline
MobileNetV2 & 0.72 & 0.55 & 18.34 & 77.87 \\
\hline
MobileNetV2+MobileNetV2 LSTM(Fusion) & 0.82 & 0.58 & 8.04 & 52.33 \\
\hline
VGG16 & 0.78 & 0.48 & 13.79 & 58.04 \\
\hline
Resnet 50 & 0.10 & 0.39 & 17.62 & 60.82 \\
\hline
MobileNetV2+VGG16 (Fusion) & 0.26 & 0.39 & 19.81 & 61.00 \\
\hline
MobileNetV2+Resnet50 (Fusion) & 0.65 & 0.32 & 23.87 & 74.69 \\
\hline
Capsule Networks & 0.04 & 0.22 & 17.87 & 68.70 \\
\hline
MobileNetV2+Capsule Networks & 0.12 & 0.22 & 13.91 & 71.32 \\
\hline
DeiT Transformer & 0.92 & 0.66 & 18.82 & 67.43 \\
\hline
MobileNetV2+DieT Transformer(Fusion) & 0.97 & 0.67 & 5.96 & 44.68 \\
\hline
\end{tabular}
\end{table*}

The MobileNetV2 model had achieved a vegetable F1 score of 0.72 and a spoilage F1 score 
of 0.55, performing moderately well on the noisy dataset. Though the noise weakens the 
ability of the model to detect tiny variations in the images, the baseline model had tried to 
identify the type of vegetable to some extent. Compared to the baseline model, the fusion 
model of MobileNetV2 and MobileNetV2-LSTM had improved its performance under 
noise, achieved a vegetable F1 score of 0.82 and a spoilage F1 score of 0.58. The regression 
metrics, MSE of 8.04 and SMAPE of 52.33\% had also decreased, indicating a strong 
performance against noise. The combination of MobileNetV2 and LSTM had improved the 
model's performance by handling both classification and regression tasks. VGG16 had 
performed moderately on the noisy dataset. All the metrics had shown moderate results, 
such as a vegetable F1 score of 0.78, a spoilage F1 score of 0.48, because the VGG16 
struggles with the texture when the noise is present. ResNet50 has been affected by noise 
by dropping the vegetable f1 score and spoilage F1 score to 0.10 and 0.39, respectively. 
Though the ResNet50 is deep, it is sensitive to the noise.

The fusion model of MobileNetv2 and VGG16 performed poorly under noise, with the vegetable F1 score and spoilage F1 score of 0.26 and 0.39, respectively. The fusion is completely failed in this combination as the VGG16 is sensitive to noise. There is no much improvement even in the regression metrics, having an MSE of 19.81 and an SMAPE of 61.00\%. The fusion model of combining MobileNetV2 and ResNet50 performed well compared with the ResNet50 base model and the VGG16 and MobileNetV3 fusion model. It had achieved the vegetable F1 score of 0.65, and the spoilage F1 score remains low with 0.32. The capsule networks had performed poorly under noisy conditions, with a very low vegetable F1 score of 0.04 and a spoilage F1 score of 0.22. This is because the noise is disrupting the capsule from forming consistent spatial relationships. The regression metrics, MSE with 17.87 and SMAPE with 68.70, also represent the inability of the model to capture the numerical patterns.

The fusion model of capsule networks has also performed poorly with a vegetable F1 score of 0.12 and a spoilage F1 score of 0.22, as the capsule networks is prone to error while noise is introduced. The error metrics are MSE of 13.91 and SMAPE of 71.32, which shows a small improvement over the baseline model but remain weak overall. The DeiT transformer shows a strong performance even on the noisy images, with an F1 score of 0.92 for vegetable and 0.66 for spoilage. These are robust to noise, as these transformers rely on global attention rather than pixel-level features. Comparatively, the classification had worked well rather than the regression because the error rate is high with MSE of 18.82 and a SMAPE of 67.43\%. The fusion model of MobileNetV2 and Deit Transformer provides the best performance in all the metrics, showing its robustness towards the noise. The vegetable F1 score reaches to 0.97, and the spoilage F1 score of 0.67 and the regression metrics shows major improvements by reducing the error rate to 5.96 and SMAPE of 44.68\%. The combination benefits using the strength of both the models capturing the local spatial features and also capturing the numerical patterns instead of using MobileNetV2 and DeiT transformer individually.

\subsection{Comparing the performance of models with the noisy images and the original dataset}

\begin{table*}[t]
\centering
\caption{Performance difference comparison of models between original and noisy datasets}
\label{tab:performance_difference}
\renewcommand{\arraystretch}{1.4}
\begin{tabular}{|p{3cm}|c|c|c|c|}
\hline
\textbf{Model} & \textbf{Vegetable Classification} & \textbf{Spoilage Detection} & \textbf{Shelf life prediction} & \textbf{Shelf life prediction} \\
& \textbf{F1 score} & \textbf{F1 score} & \textbf{MSE } & \textbf{SMAPE(\%)} \\
& \textbf{difference} & \textbf{Spoilage F1 score difference} & \textbf{difference} & \textbf{difference}\\
\hline
MobileNetV2 & 0.12 & 0.04 & 14.36 & 9.03 \\
\hline
MobileNetV2 & 0.15 & 0 & -2.97 & -9.63 \\
+MobileNetV2-LSTM(fusion) & & & & \\
\hline
VGG16 & -0.11 & 0.02 & 0.7 & 7.34 \\
\hline
ResNet50 & -0.03 & -0.05 & 0.42 & -29.22 \\
\hline
MobileNetV2& 0.15 & -0.03 & -1.37 & 2.75 \\
+VGG16(Fusion)  & & & &  \\
\hline
MobileNetV2 & 0.22 & 0.32 & 7.87 & 7.9 \\
+ResNet50(Fusion) & & & &    \\
\hline
Capsule Networks & 0 & 0.01 & 4.85 & -0.36 \\
\hline
MobileNetV2 +Capsule Networks & -0.05 & 0 & 1.34 & 7.39 \\
\hline
DeiT Transformer & 0.06 & 0.01 & -0.15 & -2.03 \\
\hline
MobileNetV2+DeiT Transformer(Fusion) & 0.01 & -0.06 & -2.38 & -3.02 \\
\hline
\end{tabular}
\end{table*}

In the vegetable F1 score differences, a positive value indicates that the model performed well on the original dataset, whereas a negative value indicates that the model had performed better on noisy images. The models like MobileNetV2, fusion model of MobileNetV2 and VGG16 and the fusion model of MobileNetV2 and ResNet50 had shown the positive differences, which indicates that the models had reduced their performance with noisy images. A few models like VGG16, ResNet50 and the fusion model of MobileNetV2 and capsule networks had performed somewhat better on the noisy dataset comparatively. The capsule networks model performed equally on the original dataset and the noisy dataset, stating that it is not affected by the noise. The transformer model had shown a difference of 0.6, whereas the fusion model had a very small difference of 0.01, stating that this model is even robust under noisy conditions.

The spoilage F1 score even follows the same interpretation as the vegetable F1 score, that the positive value represents models that perform well on the original dataset, whereas the negative value represents models that perform well on the noisy images. Some models had shown very less variations, such as Capsule networks, DeiT transformer, and MobileNetV2 and\_LSTM fusion, indicating their robustness towards the noise. The fusion model of MobileNetV2 and ResNet50 has the highest difference stating that its performance had decreased over noisy images. Some models like MobileNetV2 and Deit fusion and ResNet50 models had performed slightly better on noisy dataset.

In the MSE difference, a positive value means the model performed well on the noisy dataset, while a negative value represents the model performed well on the original dataset. Several models such as MobileNetV2, capsule networks and fusion model of MobileNetv2 and ResNet50 shows positive differences, suggesting that the models performed well on the noisy dataset. Whereas the fusion models like MobileNetV2-LSTM, MobileNetV2 VGG16, MobileNetV2-DeiT had performed less on the noisy dataset. Overall, the mixed pattern across the models highlight that some CNN models are robust to noise and some models performance had decreased form the original dataset.
\section{Conclusions}
This study successfully demonstrates that multi-task learning with fusion architectures 
represents a powerful and practical approach to comprehensive vegetable quality 
assessment that outperforms single-task and single-architecture baselines and maintains its 
robustness under realistic conditions of noise. Within the project framework, a wide array 
of deep learning models was developed and assessed, starting from traditional CNNs and 
finishing with modern transformers and novel fusion architectures, all adapted for 
simultaneous vegetable classification, spoilage detection, and shelf life prediction within a 
unified framework. The results clearly indicate that the performance of fusion 
architectures, mainly the combination of CNN-LSTM and CNN-DeiT Transformer, 
outranks the performance of their stand-alone counterparts due to the complementary 
strengths of different neural network paradigms, while CNN-LSTM achieved an 
outstanding F1 of 0.97 in vegetable classification, with very good regression capabilities 
of 5.07 MSE and 42.70\% SMAPE, and CNN-DeiT Transformer demonstrated the best 
overall performance with very good regression accuracy of 3.58 MSE and 41.66\% SMAPE 
and near-perfect vegetable classification with an F1 score of 0.98.

The robustness analysis provides crucial insights into practical deployment: model performance on clean data alone is not sufficient to predict reliability in the real world. The 
proposed CNN-DeiT Transformer fusion architecture far outperforms others on noise 
corruption, with near-optimal performance showing only minimal degradation (vegetable 
classification F1 of 0.97, MSE of 5.96, SMAPE of 44.68\% on noisy data). The reasons for 
this robustness lie in the fact that the synergistic effect of local feature extraction by CNN 
can capture noise-free image regions and extract meaningful patterns despite corruption, 
while the transformer integrates information over the entire image to average out the effects 
of noise and maintains attention on consistent features. These results establish that fusion 
architectures improve not just average performance but also provide fundamental 
robustness advantages necessary for reliable real-world deployment.

Explainability analysis using LIME gives insight into model decision-making in an 
interpretable way that instills trust and allows for the validation of learned representations. 
The generated LIME visualizations confirm that both the CNN-LSTM and CNN-DeiT 
Transformer fusion model focus on sensible, relevant features such as discoloration, 
changes in texture, and structural patterns, which human experts would consider in 
assessing produce quality rather than exploiting spurious correlations or dataset artifacts. 
The CNN-LSTM model has demonstrated more temporally coherent attention patterns 
focused on features correlating with the progression of deterioration, whereas the CNN-DeiT 
model shows sophisticated spatial reasoning that considers relationships among 
multiple regions and integrates global context to explain their respective strengths for 
different task types.

The real-world feasibility of the proposed approach is demonstrated through a successful 
deployment of best-performing models through a Flask web application, providing an end
to-end system that can immediately be adopted for real-world vegetable quality monitoring 
in supply chain management, retail operations, or automated sorting facilities. This project 
confirms the multi-task learning paradigm as an efficient approach that reduces 
computational overhead and deployment costs compared to separate models tackling each 
task individually while actually boosting performance by knowledge sharing across related 
tasks. In particular, the results of this study conclusively establish that strategic fusion of 
complementary neural network architectures—CNNs, combining spatial feature extraction 
with LSTMs for temporal modeling or transformers for global context integration—
represents a superior paradigm toward developing robust, accurate, and interpretable 
systems for agriculture and food quality applications, whose impact spreads beyond 
vegetable quality assessment to generic multi-objective optimization problems 
encompassing real-world reliability.


\begin{thebibliography}{99}

\bibitem{ref1}
Priyanka Kanupuru and N. V. Uma Reddy,
``A Deep Learning Approach to Detect the Spoiled Fruits,''
\textit{Research Gate},
vol. 10, 2022.

\bibitem{ref2}
M. Mukhiddinov, A. Muminov, and J. Cho,
``Improved Classification Approach for Fruits and Vegetables Freshness Based on Deep Learning,''
\textit{Sensors},
vol. 22, no. 21, p. 8192, 2022,
doi: 10.3390/s22218192.

\bibitem{ref3}
U. Amin, M. I. Shahzad, A. Shahzad, M. Shahzad, U. Khan, and Z. Mahmood,
``Automatic Fruits Freshness Classification Using CNN and Transfer Learning,''
\textit{Applied Sciences},
vol. 13, no. 14, p. 8087, 2023,
doi: 10.3390/app13148087.

\bibitem{ref4}
K. A. Nfor, T. P. T. Armand, K. P. I. Ismaylovna, M.-I. Joo, and H.-C. Kim,
``An Explainable CNN and Vision Transformer-Based Approach for Real-Time Food Recognition,''
\textit{Nutrients},
vol. 17, no. 2, p. 362, 2025,
doi: 10.3390/nu17020362.

\bibitem{ref5}
V. Bhole and A. Kumar,
``A Transfer Learning-Based Approach to Predict the Shelf Life of Fruit,''
\textit{Inteligencia Artificial},
vol. 24, no. 67, pp. 102--120, 2021,
doi: 10.4114/intartif.vol24iss67pp102-120.

\bibitem{ref6}
M. Iqbal, S. T. Haider, R. S. Shoukat, S. U. Rehman, K. Mahmood, S. G. Villar, L. A. D. L. Lopez, and I. Ashraf,
``Canned Apple Fruit Freshness Detection Using Hybrid Convolutional Neural Network and Transfer Learning,''
\textit{Journal of Food Quality},
vol. 2025, no. 1, Article ID 8522400, Apr. 2025,
doi: 10.1155/jfq/8522400.

\bibitem{ref7}
J. Kang and J. Gwak,
``Ensemble of multi-task deep convolutional neural networks using transfer learning for fruit freshness classification,''
\textit{Multimedia Tools and Applications},
vol. 81, pp. 22355--22377, 2022, published Aug. 24, 2021.

\bibitem{ref8}
S. S. Reka, A. Bagelikar, P. Venugopal, V. Ravi, and H. Devarajan,
``Deep learning-based classification of rotten fruits and identification of shelf life,''
\textit{Computers, Materials \& Continua},
vol. 78, no. 1, pp. 781--794, Jan. 2024,
doi: 10.32604/cmc.2023.043369.

\bibitem{ref9}
Y. Zhang, X. Yang, Y. Cheng, X. Wu, X. Sun, R. Hou, and H. Wang,
``Fruit freshness detection based on multi-task convolutional neural network,''
\textit{Current Research in Food Science},
vol. 7, p. 100733, 2024,
doi: 10.1016/j.crfs.2024.100733.

\bibitem{ref10}
Z. A. Khan, M. Waqar, K. M. Cheema, A. A. B. Mahmood, Q. Ain, N. I. Chaudhary, A. Alshehri, S. S. Alshamrani, and M. A. Z. Raja,
``EA-CNN: Enhanced attention-CNN with explainable AI for fruit and vegetable classification,''
\textit{Heliyon},
vol. 10, no. 23, p. e40820, Dec. 2024,
doi: 10.1016/j.heliyon.2024.e40820.

\bibitem{ref11}
M. A. Sofian, A. A. Putri, I. S. Edbert, and A. Aulia,
``AI-Based Recognition of Fruit and Vegetable Spoilage: Towards Household Food Waste Reduction,''
\textit{Procedia Computer Science},
vol. 245, pp. 1020--1029, 2024,
doi: 10.1016/j.procs.2024.10.330.

\bibitem{ref12}
D. Patel,
``Image Classification of Fresh and Rotten Produce Using Deep Learning,''
\textit{National High School Journal of Science},
pp. 1--10, 2023.

\bibitem{ref13}
Y. Fu, M. Nguyen, and W. Q. Yan,
``Grading Methods for Fruit Freshness Based on Deep Learning,''
\textit{SN Computer Science},
vol. 3, no. 4, p. 264, 2022,
doi: 10.1007/s42979-022-01152-7.

\bibitem{ref14}
Y. Shu, J. Zhang, Y. Wang, and Y. Wei,
``Fruit Freshness Classification and Detection Based on the ResNet-101 Network and Non-Local Attention Mechanism,''
\textit{Foods},
vol. 14, no. 11, art. no. 1987, 2025,
doi: 10.3390/foods14111987.

\bibitem{ref15}
N. Begum and M. K. Hazarika,
``Spoilage detection of tomatoes using convolutional neural network,''
\textit{Research in Agricultural Engineering},
vol. 71, pp. 80--87, 2025.

\bibitem{ref16}
L. G. Fahad, S. F. Tahir, U. Rasheed, H. Saqib, M. Hassan, and H. Alquhayz,
``Fruits and Vegetables Freshness Categorization Using Deep Learning,''
\textit{Computers, Materials \& Continua},
vol. 71, no. 3, pp. 5083--5098, 2022,
doi: 10.32604/cmc.2022.023357.

\bibitem{ref17}
Y. Yuan and X. Chen,
``Freshness detection of vegetables and fruits using deep feature fusion,''
\textit{Smart Agricultural Technology},
vol. 5, p. 100224, 2023,
doi: 10.1016/j.atech.2023.100224.

\bibitem{ref18}
A. Kazi and S. P. Panda,
``Fruit freshness classification using transfer learning with convolutional neural networks,''
\textit{Multimedia Tools and Applications},
vol. 81, no. 26, pp. 37015--37032, 2022,
doi: 10.1007/s11042-022-12150-5.

\bibitem{ref19}
E. Tapia-Mendez, I. A. Cruz-Albarran, S. Tovar-Arriaga, and L. A. Morales-Hernandez,
``Deep learning--based intelligent system for fruit and vegetable classification and ripeness evaluation,''
\textit{Applied Sciences},
vol. 13, no. 22, art. no. 12504, 2023,
doi: 10.3390/app132212504.

\bibitem{ref20}
K. Roy, S. S. Chaudhuri, and S. Pramanik,
``Detection and segmentation of rotten and fresh apples using enhanced U-Net architecture,''
\textit{Machine Vision and Applications},
vol. 31, no. 8, pp. 1--14, 2020,
doi: 10.1007/s00542-020-05123-x.

\bibitem{ref21}
M.-C. Chen, Y.-T. Cheng, and C.-Y. Liu,
``Real-time fruit quality classification using YOLO-based detection and tracking,''
\textit{Sensors},
vol. 20, no. 5, pp. 1--16, 2020.

\bibitem{ref22}
A. Bhargava and A. Bansal,
``Quality evaluation of fresh and rotten apples using image processing and machine learning techniques,''
\textit{Journal of Food Measurement and Characterization},
vol. 15, no. 4, pp. 3570--3584, 2021,
doi: 10.1007/s12161-021-01970-0.

\bibitem{ref23}
K. Oliullah, M. R. Islam, J. I. Babar, M. A. N. Quraishi, M. M. Rahman, M. Mahbub-Or-Rashid, and T. M. A.-U.-H. Bhuiyan,
``IoT-enabled hybrid deep learning approach for fresh fruit and vegetable identification,''
\textit{Results in Engineering},
vol. 22, p. 100660, 2024,
doi: 10.1016/j.rineng.2024.100660.

\bibitem{ref24}
Y. Yuan, J. Chen, K. Polat, and A. Alhudhaif,
``Fruit and vegetable freshness detection using CNN--BiLSTM fusion model,''
\textit{Smart Agricultural Technology},
vol. 6, p. 100492, 2024,
doi: 10.1016/j.atech.2024.100492.

\bibitem{ref25}
S. Joseelan D,
``FreshCheck: A Two-Stage Deep Learning Framework for Automated Fresh Produce Assessment,''
\textit{TIJER -- International Research Journal},
2025.

\bibitem{ref26}
M. Sandler, A. Howard, M. Zhu, A. Zhmoginov, and L.-C. Chen,
``MobileNetV2: Inverted Residuals and Linear Bottlenecks,''
in \textit{Proc. IEEE/CVF Conf. Computer Vision and Pattern Recognition (CVPR)},
Salt Lake City, UT, USA, pp. 4510--4520, 2018,
doi: 10.1109/CVPR.2018.00474.

\bibitem{ref27}
M. G. Sherina, R. H. Kumar, S. Harish, and R. Harikrishnan,
``Enhancing deepfake detection through hybrid MobileNet--LSTM model with real-time image and video analysis,''
\textit{International Research Journal of Modernization in Engineering Technology and Science},
vol. 7, no. 5, May 2025, e-ISSN: 2582-5208.


\bibitem{ref28}
K. Simonyan and A. Zisserman,
``Very Deep Convolutional Networks for Large-Scale Image Recognition,''
arXiv preprint arXiv:1409.1556, 2014.

\bibitem{ref29}
K. He, X. Zhang, S. Ren, and J. Sun,
``Deep Residual Learning for Image Recognition,''
in \textit{Proc. IEEE Conf. Computer Vision and Pattern Recognition (CVPR)},
pp. 770--778, 2016,
doi: 10.1109/CVPR.2016.90.

\bibitem{ref30}
T. Mim, M. M. Rahman, J. Biswas, A. Shafkat, and K. M. M. Uddin,
``FruitsMultiNet: A deep neural network approach to identify fruits through multi-scale feature fusion using mobile interface,''
\textit{Journal of Agriculture and Food Research},
vol. 22, p. 102083, Aug. 2025,
doi: 10.1016/j.jafr.2025.102083.

\bibitem{ref31}
J. Sharma, A. A. Al-Huqail, A. Almogren, and H. Doshi,
``Deep learning based ensemble model for accurate tomato leaf disease classification by leveraging ResNet50 and MobileNetV2 architectures,''
\textit{Scientific Reports},
vol. 15, no. 1, Apr. 2025,
doi: 10.1038/s41598-025-98015-x.

\bibitem{ref32}
S. Sabour, N. Frosst, and G. E. Hinton,
``Dynamic Routing Between Capsules,''
arXiv preprint arXiv:1710.09829, 2017,
doi: 10.48550/arXiv.1710.09829.

\bibitem{ref33}
J. Zhang, X. Yu, X. Lei, and C. Wu,
``A novel Capsule Network neural network based on MobileNetV2 structure for robot image classification,''
\textit{Frontiers in Neurorobotics},
vol. 16, 2022,
doi: 10.3389/fnbot.2022.1007939.


\bibitem{ref34}
H. Touvron, M. Cord, M. Douze, F. Massa, A. Sablayrolles, and H. J\'egou,
``Training data-efficient image transformers \& distillation through attention,''
arXiv preprint arXiv:2012.12877, 2020.

\bibitem{ref35}
X. Cheng, F. Lu, and Y. Liu,
``Lightweight hybrid model based on MobileNet-V2 and Vision Transformer for human–robot interaction,''
\textit{Engineering Applications of Artificial Intelligence},
vol. 127, Part B, p. 107288, Jan. 2024,
doi: 10.1016/j.engappai.2023.107288.


\end{thebibliography}
\end{document}